%% file: 01_main.tex
\documentclass{article}

\usepackage[final]{neurips_2021}

\usepackage[utf8]{inputenc} % allow utf-8 input
\usepackage[T1]{fontenc}    % use 8-bit T1 fonts
\usepackage[hidelinks]{hyperref}
\usepackage{url}            % simple URL typesetting
\usepackage{booktabs}       % professional-quality tables
\usepackage{amsfonts}       % blackboard math symbols
\usepackage{nicefrac}       % compact symbols for 1/2, etc.
\usepackage{microtype}      % microtypography
\usepackage{xcolor}         % colors
\usepackage{graphicx}
\usepackage{bm}
\usepackage{derivative}
\usepackage{subcaption}
\usepackage{tikzscale}
\usepackage{tabularx}
\usepackage{pgfplots}
\pgfplotsset{compat=1.12,
            label style={font=\scriptsize},
            tick label style={font=\tiny},  }
\usetikzlibrary{pgfplots.groupplots}
\usetikzlibrary{patterns}
\input{acronyms}

\title{Latent Task-Specific Graph Network Simulators}

\author{Philipp Dahlinger$^1$ \thanks{correspondence to \texttt{philipp.dahlinger@kit.edu}} ~~ 
Niklas Freymuth$^1$ ~ Michael Volpp$^2$\\ \textbf{Tai Hoang}$^1$ ~ \textbf{Gerhard Neumann}$^1$  \\
$^1$\normalfont{Autonomous Learning Robots,
Karlsruhe Institute of Technology,
Karlsruhe}\\
$^2$\normalfont{Bosch Center for Artificial Intelligence, Renningen, Germany}}

\usepackage{xcolor}  % colored todos

\begin{document}

\maketitle

\begin{abstract}
Simulating dynamic physical interactions is a critical challenge across multiple scientific domains, with applications ranging from robotics to material science. 
For mesh-based simulations, Graph Network Simulators (GNSs) pose an efficient alternative to traditional physics-based simulators. 
Their inherent differentiability and speed make them particularly well-suited for inverse design problems.
Yet, adapting to new tasks from limited available data is an important aspect for real-world applications that current methods struggle with.
We frame mesh-based simulation as a meta-learning problem and use a recent Bayesian meta-learning method to improve GNSs adaptability to new scenarios by leveraging context data and handling uncertainties.
Our approach, latent task-specific graph network simulator, uses non-amortized task posterior approximations to sample latent descriptions of unknown system properties. 
Additionally, we leverage movement primitives for efficient full trajectory prediction, effectively addressing the issue of accumulating errors encountered by previous auto-regressive methods. 
We validate the effectiveness of our approach through various experiments, performing on par with or better than established baseline methods. 
Movement primitives further allow us to accommodate various types of context data, as demonstrated through the utilization of point clouds during inference.
By combining GNSs with meta-learning, we bring them closer to real-world applicability, particularly in scenarios with smaller datasets.
\end{abstract}

% Keywords: Graph Network Simulators, Meta-Learning, Motion Primitives, Likelihood Computation, Task Generalization, Gaussian Mixture Model Neural Process, Probabilistic Dynamic Movement Primitives.

\input{main/1_introduction}

\input{main/2_related_work}
\input{main/4_main}
\input{main/5_experiments}
\input{main/7_conclusion}

\section{Acknowledgments}
This work was funded by the Deutsche Forschungsgemeinschaft (DFG, German Research Foundation) -- Project 5339.
This work was supported by funding from the pilot program Core Informatics of the Helmholtz
Association (HGF). The authors acknowledge support by the state of Baden-Württemberg through
bwHPC, as well as the HoreKa supercomputer funded by the Ministry of Science, Research and the
Arts Baden-Württemberg and by the German Federal Ministry of Education and Research.

\bibliography{iclr2024_conference}
\bibliographystyle{iclr2024_conference}

\appendix

\newpage

\input{appendix/experimental_protocol}
\input{appendix/additional_results}

\end{document}

%% file: acronyms.tex
\usepackage[acronym,nohypertypes={acronym,notation}]{glossaries}
\newacronym{gns}{GNS}{Graph Network Simulator}
\newacronym{mgn}{MGN}{MeshGraphNet}
\newacronym{mpc}{MPC}{Model Predictive Control}
\newacronym{mbrl}{MBRL}{Model-Based Reinforcement Learning}
\newacronym{gnn}{GNN}{Graph Neural Network}
\newacronym{sofa}{SOFA}{Simulation Open Framework Architecture}
\newacronym{mpn}{MPN}{Message Passing Network}
\newacronym{mlp}{MLP}{Multilayer Perceptron}
\newacronym{cnn}{CNN}{Convoluational Neural Network}
\newacronym{mse}{MSE}{Mean Squared Error}
\newacronym{iou}{IoU}{Intersection over Union}
\newacronym{ggns}{GGNS}{Grounding Graph Network Simulator}
\newacronym{ltsgns}{LTSGNS}{Latent Task-Specific Graph Network Simulator}
\newacronym{gmmnp}{GMM-NP}{Gaussian Mixture Model Neural Process}
\newacronym{prodmp}{ProDMP}{Probabilistic Dynamic Movement Primitives}
\newacronym{trngvi}{TRNG-VI}{Trust Region Natural Gradient Variational Inference}
\newacronym{gmm}{GMM}{Gaussian Mixture Model}
\newacronym{mp}{MP}{Movement Primitive}

%% file: main/1_introduction.tex
\section{Introduction}

%\paragraph{Simulation is important}
Simulating physical systems is a fundamental challenge across a variety of scientific fields, with applications ranging from structural mechanics~\citep{yazid2009state, zienkiewicz2005finite, stanova2015finite} over fluid dynamic~\citep{chung1978finite, zienkiewicz2013finite, connor2013finite} to electromagnetism~\citep{jin2015finite, polycarpou2022introduction, reddy1994finite}.
Mesh-based simulations are often chosen for these tasks due to the computational efficiency and accuracy of the underlying finite element method~\citep{brenner2008mathematical, reddy2019introduction}.
Yet, the diversity of modeled problems usually necessitates the development of task-specific simulators to accurately capture the relevant physical quantities~\citep{reddy2010finite}. 
Despite these efforts, these specialized simulators can be slow and cumbersome, especially for larger simulations~\citep{paszynski2016fast, hughes2005isogeometric}.

% \paragraph{\glspl{gns} are a thing and why are they cool}
As a result, data-driven models have gained traction as an appealing alternative~\citep{guo2016convolutional, da2021deep, li2022machine}. Among these, general-purpose~\glspl{gns} have become increasingly popular~\citep{battaglia2018relational, pfaff2020learning, allen2022graph, allen2023learning, linkerhagner2023grounding}. Building on~\glspl{gnn}~\citep{scarselli2009the, wu2020comprehensive, bronstein2021geometric},~\glspl{gns} encode the simulated system as an interaction graph between nodes, predicting their dynamics. 
These models offer a significant speed advantage over classical simulators \citep{pfaff2020learning} while being fully differentiable, making them highly effective for applications like inverse design~\citep{allen2022graph, xu2021end}.

% ~\citep{scheikl2023lapgym}
% Simulating deformable objects is a core challenge in numerous scientific domains, with applications ranging from \todo{Cite stuff} to \todo{more stuff}.
% People commonly use mesh-based simulations for this because they have some benefits \todo{cite}.
% Different applications e.g. \todo{give examples} require task-specific simulators to model the desired quantities correctly. 
% \paragraph{\glspl{gns} are a thing and why are they cool}
% With the rise of \glspl{gnn} \citep{battaglia2018relational}, there is a lot of advancements in the creation of \glspl{gns}, a class of simulators trained from data and predicting the dynamics of the objects with a neural network. It has interesting properties for various applications \todo{Give examples and cite} as it is faster than traditional physics simulators (\todo{cite that}) and furthermore differentiable, which is useful for \gls{mbrl} applications \todo{cite}. 

% \paragraph{Problem: Problems with GNS and our approach}
\glspl{gns} are commonly trained through simple next-step supervision~\citep{battaglia2018relational, pfaff2020learning, allen2023learning}.
During inference, entire trajectories are unrolled by iteratively predicting per-node dynamics from an initial system state. 
This process is susceptible to accumulating errors over time, especially as the input distribution diverges from the training set~\citep{brandstetter2021message, han2022predicting}.
While data augmentation strategies exist to offset this issue~\citep{pfaff2020learning, brandstetter2021message}, they neither correct mistakes once they have been made nor effectively address the challenges posed by partially known initial system states~\citep{linkerhagner2023grounding}.
Moreover, these models usually require large amounts of data to train, which is an issue in real-world scenarios where data is often sparse and the need for efficient adaptation to new tasks is crucial~\citep{linkerhagner2023grounding}. 
% data-intensive and struggle with generalization to new dynamics or object properties~\citep{todo}. 
% These issues are amplified in 
In this work, we reformulate learned mesh-based simulation as a trajectory-level meta-learning problem that uses mesh states as a context set to address these limitations.
We employ a Bayesian meta-learning approach based on non-amortized task posterior approximations~\citep{Volpp2021BayesianCA, volpp2023} for rapid adaptation to new task properties and uncertainties. 
We further mitigate the issue of error accumulation through the use of~\glspl{prodmp} ~\citep{schaal2006dynamic, paraschos2013probabilistic, li2023prodmp} to represent higher-order dynamics of mesh nodes on a trajectory level.
Combined, these methods allow us to model a distribution over unknown material properties and induced simulation trajectories from a given context set. 
When this context set contains sufficient information, the method is able to accurately determine specific system properties and adapt the simulation accordingly. 
We visualize an example in Figure~\ref{fig:general}.
A more detailed overview of our approach, called~\gls{ltsgns}\footnote{Code available at \url{https://github.com/PhilippDahlinger/ltsgns_ai4science}} is shown in Figure~\ref{fig:ltsgns_figure1}.

We validate the effectiveness of \gls{ltsgns} on challenging deformable object simulations, showing superior prediction quality compared to~\gls{mgn}, a state-of-the-art~\gls{gns}.
% and generalization capability.
% \todo{is the last part okay?}
Furthermore, we showcase that our method can incorporate real-world observations such as point-clouds.
This feature is particularly useful for applications dealing with sparse and incomplete data sets. 
The ability to accommodate such data types makes the model more applicable in practical scenarios and opens new avenues for research that require robust and generalizable graph network simulators.

\input{main/figure_wrappers/01_figure1}

%% file: main/figure_wrappers/01_figure1.tex
\begin{figure}[t]
    \centering
    \begin{subfigure}[b]{0.55\linewidth} % Adjust the width as needed
        \includegraphics[width=\linewidth]{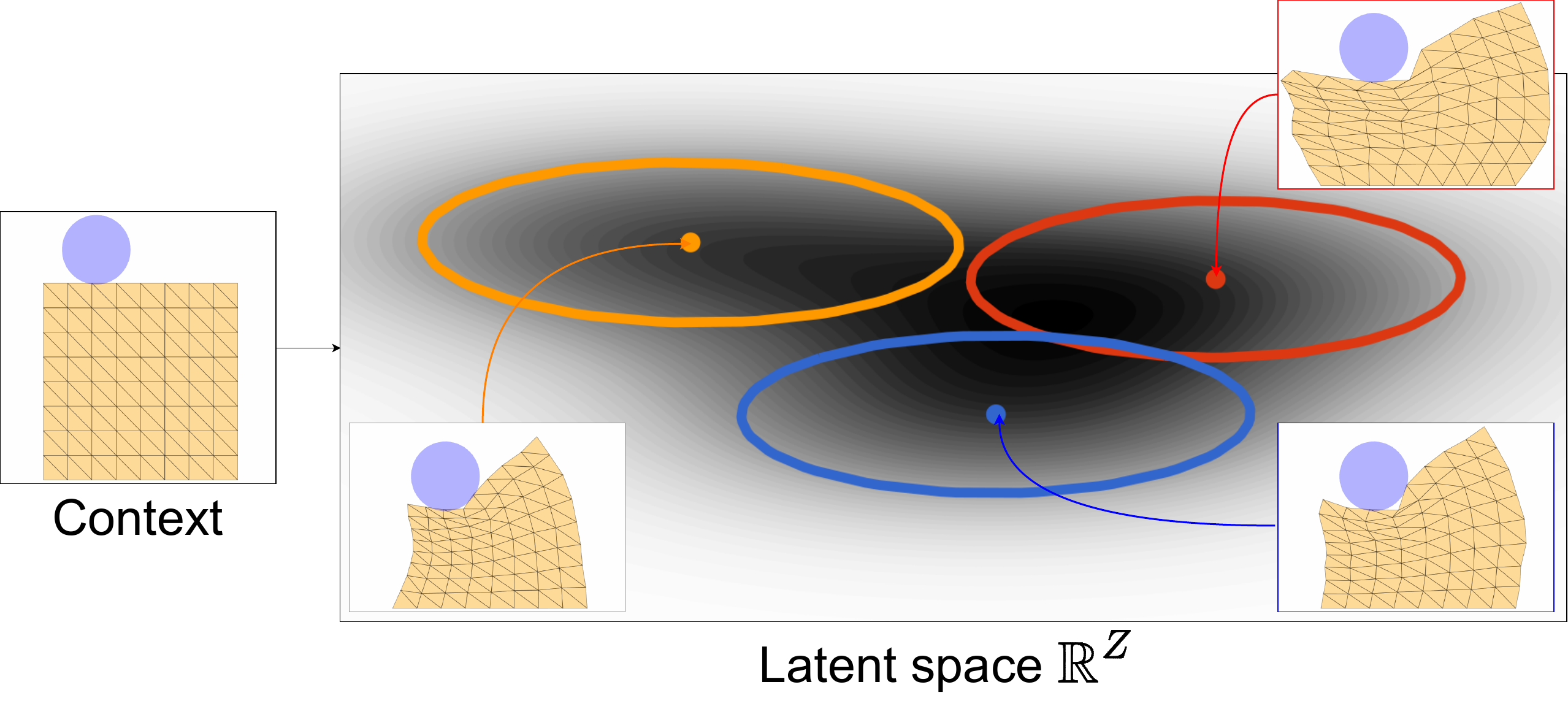} % Replace 'figure1' with your image file name
        % \caption{Subcaption for Figure 1}
        % \label{fig:sub1}
    \end{subfigure}
    \hfill % Horizontal spacing between subfigures
    \begin{subfigure}[b]{0.4\linewidth} % Adjust the width as needed
        \includegraphics[width=0.8\linewidth]{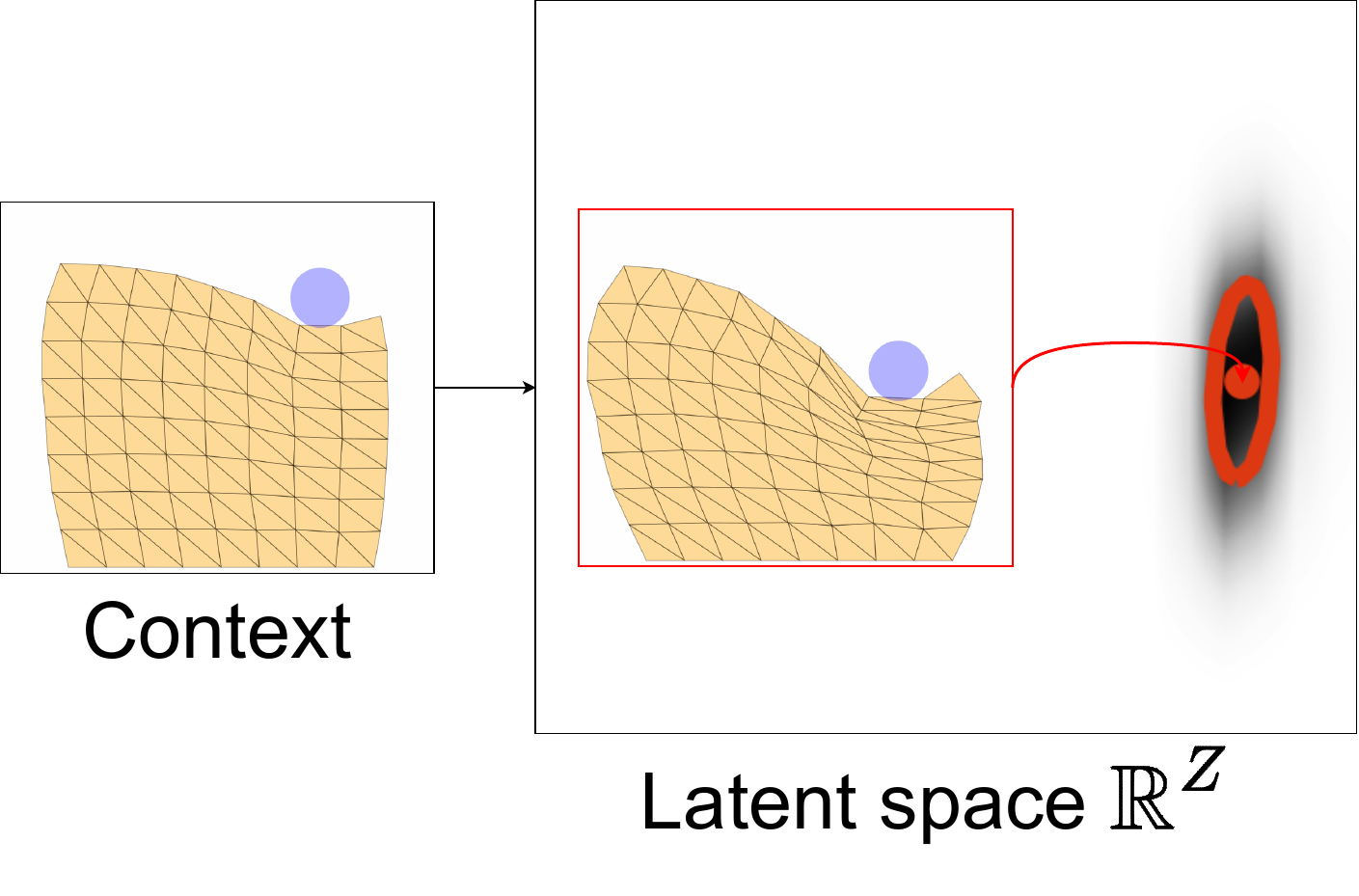} % Replace 'figure2' with your image file name
        % \caption{Subcaption for Figure 2}
        % \label{fig:sub2}
    \end{subfigure}
    
    \caption{Illustration of the latent space of \glsfirst{ltsgns} describing material properties of the mesh. 
    (Left) Given an initial mesh as a context, the task posterior depicted as a black and white contour plot is spread out to include different possible mesh deformations. 
    Thus, each sample of the posterior approximation (colored Gaussian components) results in a different yet plausible simulation outcome.
    (Right) When the material property can be inferred from additional context information, the posterior collapses to a unimodal distribution that represents deformations that are compatible with this context.
    }
    \label{fig:general}
\end{figure}

%% file: main/2_related_work.tex
\section{Related Work}
\textbf{Graph Network Simulators.} 
Recent research has increasingly focused on training deep neural networks for physical simulations as such models can yield significant speedups over traditional solvers while being fully differentiable~\citep{pfaff2020learning, allen2022physical}, making them a natural choice for e.g., model-based Reinforcement Learning~\citep{mora2021pods} and Inverse Design problems~\citep{baque2018geodesic, durasov2021debosh, allen2022physical}.
Examples use \glspl{cnn} for fluid and aerodynamic flow simulations ~\citep{tompson2016accelerating, guo2016convolutional, chu2017data-driven,zhang2018application, bhatnagar2019prediction, ummenhofer2020lagrangian,} and train either on a supervised loss or with the help of Generative Adversarial Networks~\citep{goodfellow2014generative} in an adversarial fashion~\citep{kim19deep, xie2018tempoGAN}.
A popular class of learned neural simulators are~\glspl{gns}~\citep{battaglia2016interaction, sanchezgonzalez2020learning}, a special type of~\gls{gnn}~\citep{scarselli2009the, bronstein2021geometric} that is designed to handle graph-structured physical data by modeling relations between arbitrary entities.
Here, applications include particle-based simulations~\citep{li2019learning, sanchezgonzalez2020learning}, atomic force prediction~\citep{hu2021forcenet} and fluid dynamic problems~\citep{brandstetter2021message}.
Most notably, they have been applied to the mesh-based prediction of deformable objects~\citep{pfaff2020learning, weng2021graph-based, han2022predicting, fortunato2022multiscale, linkerhagner2023grounding}.
Here, extensions include handling rigid objects~\citep{allen2022graph, allen2023learning} and integrating learned adaptive mesh refinement strategies~\citep{plewa2005adaptive, yang2023reinforcement, freymuth2023swarm} into the simulator~\citep{wu2023learning}.
Closely related to our work, another extension utilizes additional sensory information to ground simulators to improve long-term predictions~\citep{linkerhagner2023grounding}.

\paragraph{Meta Learning.}
Meta-learning~\citep{schmidhuber_learning_1992,thrun_learning_1998,vilalta_perspective_2005,Hospedales2022MetaLI} extracts inductive biases from a training set of related tasks in order to increase data efficiency on unseen tasks drawn from the same task distribution.
In contrast to other multi-task learning methods, such as transfer learning~\citep{krizhevsky2012imagenet,golovin_google_2017,Zhuang2020ACS}, which typically merely fine-tune or combine standard single-task models, meta-learning makes the multi-task setting explicit in the model architecture~\citep{bengio_learning_1991,ravi_optimization_2017,andrychowicz_learning_2016,volpp_meta-learning_2019,santoro_meta-learning_2016,snell_prototypical_2017}.
This allows the resulting meta-models to learn \textit{how} to learn new tasks with only a few context examples.
A popular variant is the model-agnostic meta-learning (MAML) family~\citep{finn_model-agnostic_2017,grant_recasting_2018,finn_probabilistic_2018,kim_bayesian_2018}, which employs standard single-task models and formulates a multi-task optimization procedure.
The neural process (NP) model family~\citep{garnelo_conditional_2018,Garnelo2018NeuralP,kim_attentive_2019,gordon_meta-learning_2019,louizos_functional_2019,Volpp2021BayesianCA} is a complementary approach in the sense that it builds on a multi-task model architecture~\citep{heskes_empirical_2000,bakker_task_2003}, but employs standard gradient based optimization algorithms~\citep{kingma_adam_2015,Kingma2014AutoEncodingVB,rezende_stochastic_2014,zaheer_deep_2017}.
Recently, \cite{volpp2023} demonstrated the importance of accurate task posterior inference for efficient meta-learning by combining an NP-like architecture with more powerful inference and optimization schemes~\citep{Arenz2018VIPS,Lin2020Handling,arenz2023}.

\paragraph{Motion Primitives.}
\glspl{mp} \citep{schaal2006dynamic, paraschos2013probabilistic} allow for compact and smooth trajectory representation via a set of basis functions.
% However, traditional approaches to fitting MPs are limited to a single task. 
Recent methods combine \glspl{mp} with neural networks to increase their expressiveness~\citep{seker2019conditional, bahl2020neural, li2023prodmp}. 
Among these, \glspl{prodmp}~\citep{li2023prodmp} introduce a novel set of basis functions that sidestep an otherwise expensive numerical integration procedure in the training pipeline while being fully differentiable.
Additionally~\glspl{prodmp} can be queried at arbitrary points in time, making them particularly suitable for our approach.

%% file: main/4_main.tex
\section{Latent Task-Specific Graph Network Simulators}
\input{main/figure_wrappers/02_figure2}

\paragraph{Graph Network Simulators}
A~\glsfirst{mpn}~\citep{sanchezgonzalez2020learning, pfaff2020learning} consists of a series of message passing steps which iteratively update latent node and edge features based on the graph topology.
Given a graph $\mathcal{G} = (\mathcal{V}, \mathcal{E}, \mathbf{X}_\mathcal{V}, \mathbf{X}_\mathcal{E})$ with nodes $\mathcal{V}$, edges $\mathcal{E}$ and associated vector-valued node and edge features $\mathbf{X}_\mathcal{V}$ and $\mathbf{X}_\mathcal{E}$, each step is given as
% Our \gls{mpn} consists of $L$ \textit{Message Passing Steps}, where each step $l$ updates latent node and edge features of a given graph using information from the previous step.
% Using \glspl{mlp} $f^l$ and initial node and edge features $\mathbf{x}_v^0$ and $\mathbf{x}_e^0$, the $l$-th step is given as
\begin{align*}
\textbf{h}^{k+1}_{e} &= f^{k}_{\mathcal{E}}(\textbf{h}^{k}_v, \textbf{h}^{k}_u, \textbf{h}^{k}_{e})\text{,}\quad
\textbf{h}^{k+1}_{v} = f^{k}_{\mathcal{V}}(\textbf{h}^{k}_{v}, \bigoplus_{e=(v,u)} \textbf{h}^{k+1}_{e})\text{,}\quad\textrm{with~} e = (u, v)\in\mathcal{E}\text{.}
\end{align*}

Here, $\textbf{h}_v^0$ and $\textbf{h}_e^0$ are embeddings of the system state per node and edge, and $\oplus$ is a permutation-invariant aggregation such as a sum, max, or mean operator. Each $f^l_\cdot$ is a learned function such as a small~\gls{mlp}.
The network's final output is a node-wise learned representation $\mathbf{h}$ that encodes local information about the graph topology and the predicted dynamics of the respective parts of the simulated system.
% This latent information can then be combined with additional information to predict the dynamics of the system
% \todo{mention that for meta-learning, this is relatively abstract?}

\glspl{gns} encode the system state as a graph, feed it through a~\gls{mpn} and interpret the outputs per node as dynamics that can be used to forward the simulation in time using e.g., a forward-Euler integrator.
The state encoding usually uses relative distances and velocities rather than absolute ones, as the resulting translation-invariance allows for better generalization over local areas~\citep{sanchezgonzalez2020learning}.
When parts of the simulation are known, such as e.g., the positions of a robot's end-effector for a planned trajectory, only the remaining nodes are predicted.
Existing \glspl{gns} usually minimize a next-step \gls{mse} per node during training and produce longer trajectories by iteratively applying the resulting forward dynamics~\citep{pfaff2020learning}.
As this iterative dependence on previous predictions causes errors to accumulate over time, carefully tuned implicit de-noising strategies are often added during training~\citep{sanchezgonzalez2020learning, pfaff2020learning, brandstetter2021message}.
% Intuitively, adding training noise acts as a form of data augmentation that allows the learned model to compensate for small prediction errors over time. 
% \todo{here: This kind of error-compensating next-step prediction leads to plausible and visually realistic predictions. However, the resulting predictions can be arbitrarily inaccurate with respect to the true dynamics of the system, since the model has no reference for its simulation other than some potentially incomplete initial state.}
Here, we instead use~\glspl{prodmp} to directly predict a compact trajectory representation per system node in a single step, reducing the effect of error accumulation similar to, e.g., temporal bundling~\citep{brandstetter2021message}.
% Previous work~\citep{pfaff2020learning} decomposes these learned functions to explicitly represent two different kinds of edges, in their case mesh and world edges.
% More concretely, they assume an edge partition $\mathbf{E} = \mathbf{E}_1 \dot{\cup} \mathbf{E}_2$ and separate edge update functions $f^l_{\mathbf{E}_1}$ and $f^l_{\mathbf{E}_2}$. 
% The edge-aggregation for the node update is then computed by aggregating the latent features of both types of edges separately and concatenating the result. 
% Our method, however, omits this explicit representation in favor of a simple one-hot encoding of the type of input edge, because we did not find any significant advantages of explicit partitioning of edge types over one-hot encodings for our tasks. For more details, see Appendix~\ref{app_sec:ablations}.
% A comparison to the explicit representation of different types of edges is conducted in our experiments.

\paragraph{Meta-Learning and Graph Network Simulators.}
We view \gls{gns} as a meta-learning problem, where each task consists of simulating a deformable object with unknown material properties over time.
% there are multiple tasks, (e.g. mesh deformations in different scenarios and environments using different material properties) 
Our goal is to learn a simulator which is adaptable to a specific scenario by observing context data. 
Following the notation of~\citet{volpp2023}, the meta-dataset $\mathcal{D} = \mathcal{D}_{1:L}$ consists of simulation \textit{tasks} $\mathcal{D}_l = \{\mathcal{G}_l, x_{l,1:T}, \bm{y}_{l,1:T}\}$ with an initial mesh $\mathcal{G}_l$ and time steps $x_{l,t}$ of the node positions $\bm{y}_{l, t} \in \mathbb{R}^{N \times D}$. We use $N$ as the number of nodes and $D$ for the world dimension of the simulation (usually $D=3$ or $D=2$). 
In contrast to standard meta-learning, we additionally use the initial graph of the system as a task-level context.
% use an additional task-level global context graph $\mathcal{G}_l$ describing the deformable mesh and a possible rigid collider. 
Given the initial positions of the deformable mesh and rigid collider, the task is to predict the node positions $\bm{y}_{l, t}$ at time steps $x_{l, t}$.
% given that the initial position is defined by $\mathcal{G}_l$. 
We note that the initial graph $\mathcal{G}_l$ does not contain the full system state, as we consider the material properties of the deformable object to be unknown.

\paragraph{Model architecture.}
Our model likelihood $L = p_{\bm \theta} (\bm y_{l, t} \mid x_{l, t}, \bm z_l, \mathcal{G}_l)$ is parametrized by a global parameter $\bm \theta \in \mathbb R^{d_{\bm \theta}}$ and defines the probability distribution over targets $\bm y_{l,t}$ at corresponding timesteps $x_{l,t}$, conditioned on a latent task descriptor $\bm z_l \in \mathbb R^Z$ and the initial graph $\mathcal{G}_l$, cf.~Fig.~ \ref{fig:ltsgns_figure1}. Given the graph $\mathcal{G}_l$, the initial mesh of a deformable object and a tractable rigid collider, we connect both objects based on proximity, compute relative distances between nodes and store this information in the edges. 
We then use a~\gls{mpn} to encode latent features $\bm h \in \mathbb{R}^H$ per node. We combine this encoding with the global latent variable $\bm z_l \in \mathbb{R}^Z \sim q_{\bm \phi_l}(\bm z_l)$ by concatenating $\bm z_l$ to every node feature $\bm h$. 
Intuitively, the latent variable $\bm z_l$ encodes the material properties and high-level deformations of the respective simulation, but does not focus on individual nodes.
We use a simple \gls{mlp} as the node-level decoder to yield final predictions per node.

% \paragraph{ProDMP usage}
Instead of iteratively predicting dynamics for the current simulation step like existing~\gls{gns}~\citep{pfaff2020learning, allen2023learning}, we use~\gls{prodmp}~\citep{li2023prodmp} to predict a representation of the full trajectory.
% One of the main difference to the \gls{mgn} architecture is the usage of movement primitives: while standard \glspl{gns} predict velocities or accelerations of the current step, we predict the  full trajectory using \glspl{prodmp} from~\citep{li2023prodmp}. 
The node-wise \gls{prodmp} weights $w \in \mathbb{R}^W$ define a trajectory over the full time horizon, and the \gls{prodmp} framework allows an efficient backpropagation from the trajectories to the weights $w$.
% Although not using distributions over trajectories, the \gls{prodmp} framework allows an efficient backpropagation from the trajectories to the weights $w$.
As such, we can predict the node positions  $\bm y_{l, t}$ at the desired time steps $x_{l, t}$ and use a node-wise Gaussian log-likelihood between the given and the predicted node positions to fit our latent space as detailed below.

\paragraph{Model predictions.}
Under our model, the predictive distribution for a target task $\mathcal D_*$, given a set of context examples \mbox{$\mathcal D^c_* \subset \mathcal D_*$},
%\todo{write better wrt $G_*$, $\mathcal D^c_*$ shall contain $\mathcal{G}_*$}
is given by
\begin{equation}
    p_{\bm{\theta}}(\bm{y}_{*,1:T} \mid  x_{*,1:T}, \mathcal D^c_*) 
    = \int \prod_{t=1}^T p_{\bm{\theta}}(\bm y_{*,t} \mid x_{*,t}, \mathcal{G}_*, \bm z_*) p_{\bm \theta}(\bm z_* \mid \mathcal D^c_*) \odif{\bm z_*},
\end{equation}
where the \textit{task posterior} distribution is given in terms of the model and a prior distribution $p(\bm z_*)$ over task descriptors by means of Bayes' theorem as
\begin{equation}
    p_\theta(\bm z_* \mid \mathcal D_*^c) 
    = \prod_{t=1}^{T^c} p_{\bm \theta} (\bm y_{*,t}^c \mid x_{*,t}^c, \bm z_*, \mathcal{G}_*) p(\bm z_*) \, / \, Z_*^c(\bm \theta) \equiv  \Tilde{p}_{\bm \theta}(\bm z_*) \, / \, Z_*^c(\bm \theta).
\end{equation}
Here, $Z_*^c(\bm \theta)$ denotes the marginal likelihood of the context data, i.e.,
\begin{equation}
    Z_*^c(\bm \theta) 
    \equiv p_{\bm{\theta}}(\bm{y}_{*,1:T^c}^c \mid x_{*,1:T^c}^c, \mathcal{G}_*) = \int \prod_{t=1}^{T^c} p_{\bm{\theta}}(\bm y_{*,t}^c \mid x_{*,t}^c, \bm z_*, \mathcal{G}_*)p(\bm z_*) \odif{\bm z_*}.
\end{equation}
For reasonably complex models, the marginal likelihood and, thus, the posterior distribution is intractable and requires approximation.
As shown by~\citet{volpp2023}, the predictive accuracy is highly dependent on the accuracy of the \textit{task posterior approximation}, causing us to mimic their approach and employ an expressive full-covariance \gls{gmm} of the form
\begin{equation}
    p_{\bm \theta}(\bm z_* \mid \mathcal D_*) 
    \approx q_{\bm \phi_*}(\bm z_*)
    = \sum_{k=1}^{K} w_{*,k} \mathcal{N}(\bm z_* \mid \bm \mu_{*,k}, \bm \Sigma_{*,k}).
\end{equation}
We use $K$ mixture components with corresponding weights $w_{*,k}$, means $\bm \mu_{*,k}$, and covariance \mbox{matrices $\bm \Sigma_{*,k}$}, which we collective denote by $\bm \phi_*$.
To fit the variational distribution $q_{\bm \phi_*}(\bm z_*)$, we also follow~\citet{volpp2023} and use the \gls{trngvi} method, specifically SEMTRUX~\citep{arenz2023}. 
This requires samples of $\nabla_z \Tilde{p}_{\bm \theta}(\bm z_*)$, which can be readily obtained using standard automatic differentiation tools~\citep{pytorch}.

\paragraph{Meta-training.}
The aim of meta-learning is to automatically encode inductive biases towards the task distribution extracted from the meta-dataset $\mathcal D$ in the task-global parameter $\bm \theta$.
To this end, we maximize w.r.t.~$\bm \theta$ the log marginal likelihood, which is given as the sum of the per-task log marginal likelihoods 
\begin{equation}
    \log Z_l(\bm \theta) 
    \equiv \log p_{\bm{\theta}}(\bm{y}_{l,1:T} \mid x_{l,1:T}, \mathcal{G}_l) 
    = \log \int \prod_{t=1}^T p_{\bm{\theta}}(\bm y_{l, t} \mid x_{l,t}, \bm z_l, \mathcal{G}_l)p(\bm z_l) \odif{\bm z_l}.
\end{equation}
As discussed above, the marginal likelihood is intractable, which is why we employ an evidence lower bound (ELBO) of the form
\begin{equation*}
    \text{ELBO}_l(\bm \theta) = \mathbb{E}_{q_{\bm \phi_l}(\bm z_l)}\left(\sum_{t=1}^T \log p_{\bm \theta}(\bm y_{l, t} \mid x_{l, t}, \bm z_l, \mathcal{G}_l) + \log \frac{p(\bm z_l)}{q_{\phi_l}(\bm z_l)}\right)
    \leq \log Z_\ell(\bm \theta)
\end{equation*}
as a surrogate objective for maximization of the log marginal likelihood.
The efficiency of this optimization scheme increases with the tightness of the lower bound, which is in turn controlled by the task posterior approximation quality of the variational distribution $q_{\bm \phi_{l}}$~\citep{bishop_pattern_2006,volpp2023}.
Therefore, we also employ the expressive GMM-TRNG-VI approximation procedure to fit $q_{\bm \phi_l}$ during meta-training.
The resulting ELBO can then be efficiently approximated using a Monte-Carlo estimation of the expectation, and be optimized using standard gradient-based optimization~\citep{kingma_auto-encoding_2013-1,rezende_stochastic_2014,volpp2023,kingma_adam_2015}.
After meta-training, we fix the parameters $\bm \theta$ and use them for predictions on unseen tasks.

%% file: main/figure_wrappers/02_figure2.tex
\begin{figure}
    \centering
	\includegraphics[width=0.9\textwidth]{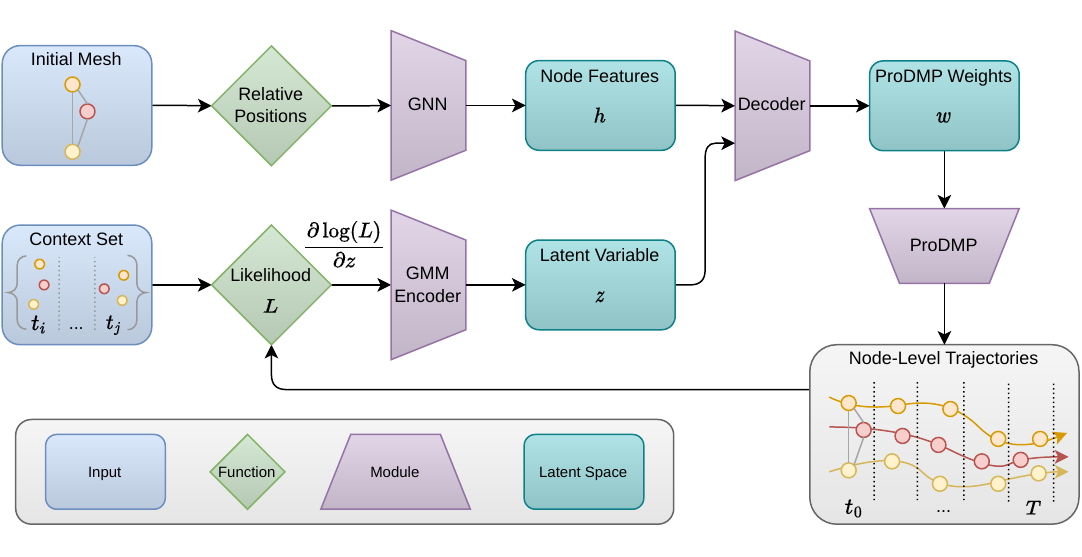}
	\caption{Schematic of the \gls{ltsgns} architecture. We compute relative node positions stored in the edges of initial mesh and obtain latent node features through a \gls{gnn}. Combining the node features with a latent task-specific variable, we compute a trajectory per node using \glspl{prodmp}. To get the latent variable $z$ for the current context, we approximate the task posterior with a \glsfirst{gmm}. This requires the gradient of the likelihood with respect to $z$. }
    \label{fig:ltsgns_figure1}
\end{figure}

%% file: main/5_experiments.tex
\section{Experiments}
\input{main/figure_wrappers/04_figure_bar_plots_deformable_plate}
\paragraph{Setup.}
We base our experimental setup on that of~\citet{linkerhagner2023grounding}, adapting it to meta-learning and the prediction of full trajectories using movement primitives.
To represent the system as a graph, we employ one-hot encoding to differentiate between deformable objects and colliders, and encode relative distances between neighboring nodes in their edges.
We do not use explicit world edges, but include the distances between nodes in mesh space.
We utilize the context set $\mathcal{D}^c_*$, comprising the initial mesh and $C$  randomly sampled simulation states, to predict node-wise \gls{prodmp} parameters $w$. These parameters encode the full mesh trajectory, and can be queried to obtain the system state at arbitrary timesteps.

All models are optimized using Adam~\citep{kingma2014adam} with a learning rate of $5\times10^{-4}$.
For the \glspl{mpn}, we use $5$ separate message passing steps for all methods.
For~\gls{ltsgns}, we repeat each step $2$ times in an inner loop to increase the receptive field of the individual nodes, as there is no iterative prediction that could otherwise pass implicit information.
Each block uses a latent dimension of $128$, LeakyReLU activations and a $1$-layer \glspl{mlp} for its node and edge updates.
We additionally apply Layer Normalization~\citep{ba2016layer} and Residual Connections~\citep{he2016deep} independently for both node and edge updates.
We repeat each experiment for $5$ random seeds and report the mean and standard deviation over these seeds. 
Each run's results are averaged over all trajectory steps and test set trajectories.
% We evaluate the models using a $k$-step prediction loss, which is the defined as the accumulated mean squared error over mesh node positions for $1$ to $k$ environment steps.
% When $k$ is set to the length of the simulation, this metric equals the rollout mean squared error.
We evaluate the models using the \textit{Rollout \gls{mse}}, which is the average~\glspl{mse} of all simulation steps, and the \textit{Last Step \gls{mse}}, which is the error of the final simulation step.
% both the cumulative mean squared prediction error over the full trajectory and the error at the final simulation step.
% of the method aa $k$-step prediction loss, which is the defined as the accumulated mean squared error over mesh node positions for $1$ to $k$ environment steps.
% When $k$ is set to the length of the simulation, this metric equals the rollout mean squared error.
% We additionally measure the training loss, i.e., the $1$-step prediction loss over random environment steps as this metric indicates the general capacity of the model to learn simulation behavior from the provided data.
Additional details on our experimental setup are provided in Appendix~\ref{app_sec:experimental_protocol}.

\paragraph{Tasks.}
We consider a simpler $2$-dimensional \textit{Deformable Plate} and a more challenging $3$-dimensional \textit{Tissue Manipulation} task~\citep{linkerhagner2023grounding}.
Both tasks use \gls{sofa}~\citep{faure2012sofa} to generate the underlying ground truth data, and use triangular surface meshes for the simulation.
While the initial meshes are known, both tasks use materials with a randomized and unknown Poisson's ratio~\citep{lim2015auxetic} that governs whether the material contracts or expands under deformation.
% This variability is especially relevant in robotics and material science, where accurate simulation of materials with diverse properties is crucial.
The \textit{Deformable Plate} task simulates different trapezoids that are deformed by a circular collider with constant velocity and varying size and starting position. 
Each trajectory consists of a mesh with $81$ nodes that is deformed over $50$ timesteps, and we use $675/135/135$ trajectories for training, evaluation and testing respectively.
The \textit{Tissue Manipulation} task simulates a common scenario in surgical robotics where a piece of tissue is deformed by a gripper. 
Here, the gripper starts attached to a random position of the object and moves in a random direction with constant velocity. The simulated mesh has $361$ nodes, and we use  $600/120/120$ training, evaluation and testing trajectories with $100$ steps each.
All tasks are normalized to be in $[-1, 1]$.

\paragraph{Baselines and Ablations.}
We use \gls{mgn}~\citep{pfaff2020learning} as our main baseline. 
\gls{mgn} iteratively predicts the velocities of the current simulation step to generate the next mesh state. 
It is trained to minimize the $1$-step \gls{mse} over node velocities and crucially employs Gaussian input noise~\citep{brandstetter2021message} to prevent error accumulation over time and thus generalize from $1$-step predictions to larger rollouts during inference.
We follow previous work~\citep{linkerhagner2023grounding} and set the standard deviation of the input noise to $0.01$.
We experiment with both \gls{mgn} without information about the material properties, and with a variant that includes this additional information, called MGN(M).
The latter renders the simulation deterministic with respect to the initial system state and sets an upper performance limit for the standard MGN model.
Additionally, we compare to an MGN(MP) variant that incorporate \glspl{prodmp} to directly predict a trajectory for each node feature instead of iteratively predicting the next state.
Since this method directly predicts the full trajectory, it does not use Gaussian input noise.
Building on our model's adaptability to different context sets, we conduct an ablation experiment focused on practical applications. 
Specifically, we test the use of point clouds as the context set during inference. 
This is particularly relevant as point clouds can be readily generated from depth cameras in real-world settings, while mesh states cannot. 
Importantly, this adjustment requires no modifications to the existing training process.

% Building on our model's capability to adapt to various context sets, we conduct an ablation experiment that considers a more practical applications. 
% Specifically, we experiment with using point clouds as the context set during inference while leaving the training untouched, as point clouds can easily be generated from depth cameras in real-world scenarios, whereas mesh states can not.
% This approach is grounded in realism, as point clouds can be derived from sensory information, thereby enabling us to model mesh movement over time. An added advantage is that our model still trains on meshes, eliminating the need for any alterations in the training process.

\input{main/figure_wrappers/05_figure_bar_plots_tissue_manipulation}
\textbf{Results.}
Figure~\ref{fig:bar_chart_deformable_plate} shows results for the \textit{Deformable Plate} task.
We find that \gls{ltsgns} outperforms~\gls{mgn} even when provided with just a single context point.
%, showcasing its robustness and ability to adapt to scenarios. 
The model's performance continues to improve as the size of the context set increases.
Specifically, for a context set with $10$ points, \gls{ltsgns} outperforms \gls{mgn}(M) even though the latter has direct access to the ground truth material information. 
Although the performance drops when using point cloud data as context instead of system states, \gls{ltsgns} still outperforms \gls{mgn} and benefits from the addition of more contextual information.
Similarly, Figure~\ref{fig:bar_chart_tissue_manipulation} evaluates the \textit{Tissue Manipulation} task. 
Here, \gls{ltsgns} again improves with an increasing context size, outperforming all~\gls{mgn} baselines for $5$ or more context points.
Providing material properties still improves~\gls{mgn}, but the difference is less significant than for the \textit{Deformable Plate} task, presumably because the dynamics are overall harder to predict even with this additional information.
We provide additional results for larger context sets in Appendix~\ref{app_ssec:additional_results_quantitative}.

\input{main/figure_wrappers/03_figure_qualitative_results}
Figure~\ref{fig:qualitative_results} visualizes exemplary final simulation steps, supporting the findings above. 
\gls{ltsgns} accurately simulates the object's deformation from a single context point, and further improves when provided with additional context information.
Opposed to this,~\gls{mgn}, even when provided with explicit material properties or when combined with~\glspl{prodmp}, fails to produce consistent meshes.
Appendix~\ref{app_ssec:additional_results_qualitative} shows visualizations for full rollouts.

%% file: main/figure_wrappers/04_figure_bar_plots_deformable_plate.tex
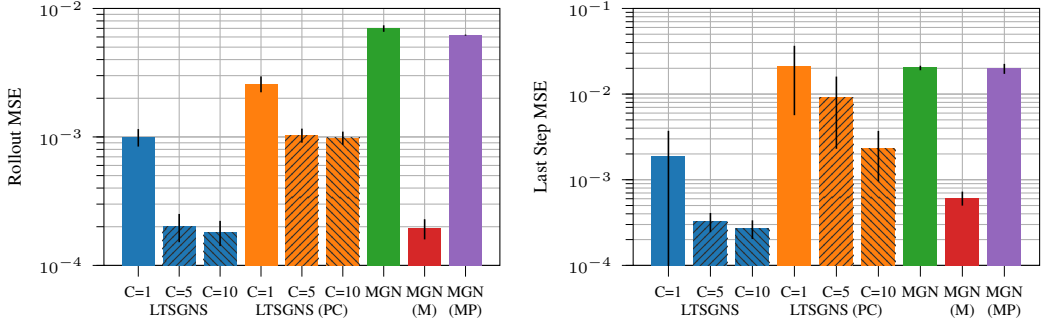
\begin{figure}[t]
    \centering
    \begin{subfigure}[b]{0.49\linewidth} % Adjust the width as needed
        \input{figures/tikz/deformable_plate}
    \end{subfigure}
    \hfill % Horizontal spacing between subfigures
    \begin{subfigure}[b]{0.5\linewidth} % Adjust the width as needed
        \input{figures/tikz/deformable_plate_last_step}
    \end{subfigure}
    \caption{
    (Left) Rollout and (Right) Last Step \gls{mse} of the different methods on the \textit{Deformable Plate} task. 
    Both metrics are plotted on a logarithmic scale.
    \gls{ltsgns} outperforms~\gls{mgn} and its \gls{mp} variant for both metrics from a single context point, further improving its performance when provided with additional context information. 
    During inference, point cloud information can be used as a context set, making for easier context acquisition at the cost of slightly worse predictions.
    Overall, the performance metrics for both Rollout and Last Step MSE are fairly similar. 
    However, the Last Step MSE exhibits increased variance across all methods, making it less consistent in comparison.
    }
    \label{fig:bar_chart_deformable_plate}
\end{figure}

%% file: figures/tikz/deformable_plate.tex
% This file was created with tikzplotlib v0.10.1.
\begin{tikzpicture}

\definecolor{crimson2143940}{RGB}{214,39,40}
\definecolor{darkgray176}{RGB}{176,176,176}
\definecolor{darkorange25512714}{RGB}{255,127,14}
\definecolor{forestgreen4416044}{RGB}{44,160,44}
\definecolor{mediumpurple148103189}{RGB}{148,103,189}
\definecolor{steelblue31119180}{RGB}{31,119,180}
\definecolor{darkslategray46}{RGB}{46,46,46}
\begin{axis}[
width=\textwidth,
height=5cm,
log basis y={10},
tick align=outside,
tick pos=left,
x grid style={darkgray176},
xmajorgrids,
xmin=-0.84, xmax=8.84,
xtick style={color=black},
xtick={0,1,2,3,4,5,6,7,8},
xticklabel style={align=center},
xticklabels={
  C=1,
  C=5 \\ \gls{ltsgns},
  C=10,
  C=1,
  C=5 \\ \gls{ltsgns} (PC),
  C=10,
  \gls{mgn},
  \gls{mgn} \\ (M),
  \gls{mgn} \\ (MP),
},
y grid style={darkgray176},
ylabel={Rollout MSE},
ymajorgrids,
ymin=0.0001, ymax=0.01,
ymode=log,
ytick style={color=black},
grid=both,
yminorticks=true,
% yticklabels={
%   \(\displaystyle {10^{-5}}\),
%   \(\displaystyle {10^{-4}}\),
%   \(\displaystyle {10^{-3}}\),
%   \(\displaystyle {10^{-2}}\),
%   \(\displaystyle {10^{-1}}\),
%   \(\displaystyle {10^{0}}\)
% }
]
\draw[draw=none,fill=steelblue31119180] (axis cs:-0.4,0.00000001) rectangle (axis cs:0.4,0.000995992333628237);
\draw[draw=none,fill=steelblue31119180, postaction={pattern=north east lines, pattern color=darkslategray46}] (axis cs:0.6,0.00000001) rectangle (axis cs:1.4,0.000201520411064848);
\draw[draw=none,fill=steelblue31119180, postaction={pattern=north west lines 	, pattern color=darkslategray46}] (axis cs:1.6,0.00000001) rectangle (axis cs:2.4,0.000181714750942774);
\draw[draw=none,fill=darkorange25512714] (axis cs:2.6,0.00000001) rectangle (axis cs:3.4,0.00259234984405339);
\draw[draw=none,fill=darkorange25512714, postaction={pattern=north east lines, pattern color=darkslategray46}] (axis cs:3.6,0.00000001) rectangle (axis cs:4.4,0.00103087805910036);
\draw[draw=none,fill=darkorange25512714, postaction={pattern=north west lines 	, pattern color=darkslategray46}] (axis cs:4.6,0.00000001) rectangle (axis cs:5.4,0.000983766315039247);
\draw[draw=none,fill=forestgreen4416044] (axis cs:5.6,0.00000001) rectangle (axis cs:6.4,0.00698697278276086);
\draw[draw=none,fill=crimson2143940] (axis cs:6.6,0.00000001) rectangle (axis cs:7.4,0.00019408235384617);
\draw[draw=none,fill=mediumpurple148103189] (axis cs:7.6,0.00000001) rectangle (axis cs:8.4,0.00620234319940209);
\path [draw=black, semithick]
(axis cs:0,0.000841293610898635)
--(axis cs:0,0.00115069105635784);

\path [draw=black, semithick]
(axis cs:1,0.000151577272235135)
--(axis cs:1,0.000251463549894561);

\path [draw=black, semithick]
(axis cs:2,0.000141369666997861)
--(axis cs:2,0.000222059834887688);

\path [draw=black, semithick]
(axis cs:3,0.00222597537485019)
--(axis cs:3,0.00295872431325658);

\path [draw=black, semithick]
(axis cs:4,0.00090185292998838)
--(axis cs:4,0.00115990318821234);

\path [draw=black, semithick]
(axis cs:5,0.000867873934405715)
--(axis cs:5,0.00109965869567278);

\path [draw=black, semithick]
(axis cs:6,0.00657996642208122)
--(axis cs:6,0.0073939791434405);

\path [draw=black, semithick]
(axis cs:7,0.000159421332010872)
--(axis cs:7,0.000228743375681468);

\path [draw=black, semithick]
(axis cs:8,0.00614262091360138)
--(axis cs:8,0.00626206548520281);

\end{axis}

\end{tikzpicture}

%% file: figures/tikz/deformable_plate_last_step.tex
% This file was created with tikzplotlib v0.10.1.
\begin{tikzpicture}

\definecolor{crimson2143940}{RGB}{214,39,40}
\definecolor{darkgray176}{RGB}{176,176,176}
\definecolor{darkorange25512714}{RGB}{255,127,14}
\definecolor{forestgreen4416044}{RGB}{44,160,44}
\definecolor{mediumpurple148103189}{RGB}{148,103,189}
\definecolor{steelblue31119180}{RGB}{31,119,180}
\definecolor{darkslategray46}{RGB}{46,46,46}

\begin{axis}[
width=\textwidth,
height=5cm,
log basis y={10},
tick align=outside,
tick pos=left,
x grid style={darkgray176},
xmajorgrids,
xmin=-0.84, xmax=8.84,
xtick style={color=black},
xtick={0,1,2,3,4,5,6,7,8},
xticklabel style={align=center},
xticklabels={
  C=1,
  C=5 \\ \gls{ltsgns},
  C=10,
  C=1,
  C=5 \\ \gls{ltsgns} (PC),
  C=10,
  \gls{mgn},
  \gls{mgn} \\ (M),
  \gls{mgn} \\ (MP),
},
y grid style={darkgray176},
ylabel={Last Step MSE},
ymajorgrids,
ymin=0.0001, ymax=0.1,
ymode=log,
ytick style={color=black},
grid=both,
yminorticks=true,
% yticklabels={
%   \(\displaystyle {10^{-6}}\),
%   \(\displaystyle {10^{-5}}\),
%   \(\displaystyle {10^{-4}}\),
%   \(\displaystyle {10^{-3}}\),
%   \(\displaystyle {10^{-2}}\),
%   \(\displaystyle {10^{-1}}\),
%   \(\displaystyle {10^{0}}\)
% }
]
\draw[draw=none,fill=steelblue31119180] (axis cs:-0.4,0.00000001) rectangle (axis cs:0.4,0.00188935299520381);
\draw[draw=none,fill=steelblue31119180, postaction={pattern=north east lines, pattern color=darkslategray46}] (axis cs:0.6,0.00000001) rectangle (axis cs:1.4,0.000326562489499338);
\draw[draw=none,fill=steelblue31119180, postaction={pattern=north west lines 	, pattern color=darkslategray46}] (axis cs:1.6,0.00000001) rectangle (axis cs:2.4,0.000269319023936987);
\draw[draw=none,fill=darkorange25512714] (axis cs:2.6,0.00000001) rectangle (axis cs:3.4,0.0211475285701454);
\draw[draw=none,fill=darkorange25512714, postaction={pattern=north east lines, pattern color=darkslategray46}] (axis cs:3.6,0.00000001) rectangle (axis cs:4.4,0.00915149240754545);
\draw[draw=none,fill=darkorange25512714, postaction={pattern=north west lines 	, pattern color=darkslategray46}] (axis cs:4.6,0.00000001) rectangle (axis cs:5.4,0.00233757894602604);
\draw[draw=none,fill=forestgreen4416044] (axis cs:5.6,0.00000001) rectangle (axis cs:6.4,0.0202072292566299);
\draw[draw=none,fill=crimson2143940] (axis cs:6.6,0.00000001) rectangle (axis cs:7.4,0.000612456636736169);
\draw[draw=none,fill=mediumpurple148103189] (axis cs:7.6,0.00000001) rectangle (axis cs:8.4,0.0198734890669584);
\path [draw=black, semithick]
(axis cs:0,5.37696451555273e-05)
--(axis cs:0,0.00372493634525209);

\path [draw=black, semithick]
(axis cs:1,0.000244089224919965)
--(axis cs:1,0.000409035754078711);

\path [draw=black, semithick]
(axis cs:2,0.000203066994238882)
--(axis cs:2,0.000335571053635092);

\path [draw=black, semithick]
(axis cs:3,0.0056708055513405)
--(axis cs:3,0.0366242515889502);

\path [draw=black, semithick]
(axis cs:4,0.00229686863880313)
--(axis cs:4,0.0160061161762878);

\path [draw=black, semithick]
(axis cs:5,0.000958810427811508)
--(axis cs:5,0.00371634746424058);

\path [draw=black, semithick]
(axis cs:6,0.0189857797911234)
--(axis cs:6,0.0214286787221365);

\path [draw=black, semithick]
(axis cs:7,0.000498838144039094)
--(axis cs:7,0.000726075129433245);

\path [draw=black, semithick]
(axis cs:8,0.017221661287112)
--(axis cs:8,0.0225253168468049);

\end{axis}

\end{tikzpicture}

%% file: main/figure_wrappers/05_figure_bar_plots_tissue_manipulation.tex
\begin{figure}[t]
    \centering
    \begin{subfigure}[b]{0.49\linewidth} % Adjust the width as needed
        \input{figures/tikz/tissue_manipulation_mesh_context}
    \end{subfigure}
    \hfill % Horizontal spacing between subfigures
    \begin{subfigure}[b]{0.5\linewidth} % Adjust the width as needed
        \input{figures/tikz/tissue_manipulation_last_step}
    \end{subfigure}
    \caption{
        (Left) Rollout and (Right) Last Step \gls{mse} of the different methods on the \textit{Tissue Manipulation} task. 
        Both metrics are plotted on a logarithmic scale.
        \gls{ltsgns} yields more accurate simulations with increasing context size, clearly outperforming the~\gls{mgn} baselines for $5$ or more context points.
    }
    \label{fig:bar_chart_tissue_manipulation}
\end{figure}
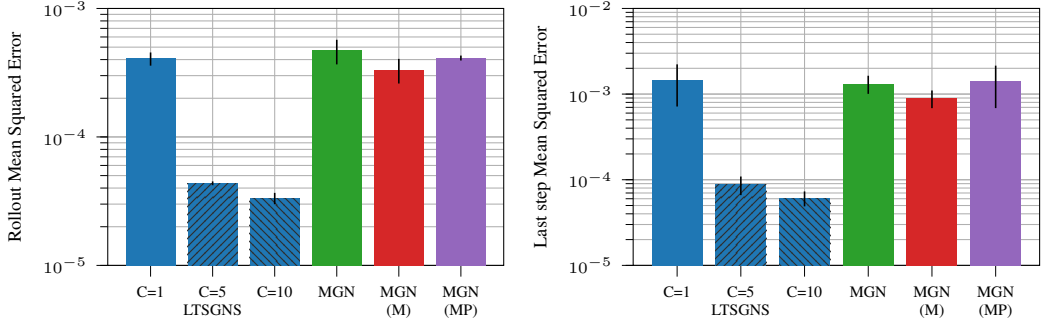

%% file: figures/tikz/tissue_manipulation_mesh_context.tex
% This file was created with tikzplotlib v0.10.1.
\begin{tikzpicture}

\definecolor{crimson2143940}{RGB}{214,39,40}
\definecolor{darkgray176}{RGB}{176,176,176}
\definecolor{forestgreen4416044}{RGB}{44,160,44}
\definecolor{mediumpurple148103189}{RGB}{148,103,189}
\definecolor{steelblue31119180}{RGB}{31,119,180}
\definecolor{darkslategray46}{RGB}{46,46,46}

\begin{axis}[
width=\textwidth,
height=5cm,
log basis y={10},
tick align=outside,
tick pos=left,
x grid style={darkgray176},
xmajorgrids,
xmin=-0.69, xmax=5.69,
xtick style={color=black},
xtick={0,1,2,3,4,5},
xticklabel style={align=center},
xticklabels={
  C=1,
  C=5 \\ \gls{ltsgns},
  C=10,
  \gls{mgn},
  \gls{mgn} \\ (M),
  \gls{mgn} \\ (MP)
},
y grid style={darkgray176},
ylabel={Rollout Mean Squared Error},
ymajorgrids,
ymin=0.00001, ymax=0.001,
ymode=log,
ytick style={color=black},
grid=both,
yminorticks=true,
% yticklabels={
%   \(\displaystyle {10^{-6}}\),
%   \(\displaystyle {10^{-5}}\),
%   \(\displaystyle {10^{-4}}\),
%   \(\displaystyle {10^{-3}}\),
%   \(\displaystyle {10^{-2}}\)
% }
]
\draw[draw=none,fill=steelblue31119180] (axis cs:-0.4,0.00000001) rectangle (axis cs:0.4,0.000406562507851049);
\draw[draw=none,fill=steelblue31119180, postaction={pattern=north east lines, pattern color=darkslategray46}] (axis cs:0.6,0.00000001) rectangle (axis cs:1.4,4.35687761637382e-05);
\draw[draw=none,fill=steelblue31119180,  postaction={pattern=north west lines, pattern color=darkslategray46}] (axis cs:1.6,0.00000001) rectangle (axis cs:2.4,3.33160252921516e-05);
\draw[draw=none,fill=forestgreen4416044] (axis cs:2.6,0.00000001) rectangle (axis cs:3.4,0.000468858386739157);
\draw[draw=none,fill=crimson2143940] (axis cs:3.6,0.00000001) rectangle (axis cs:4.4,0.000332854135194793);
\draw[draw=none,fill=mediumpurple148103189] (axis cs:4.6,0.00000001) rectangle (axis cs:5.4,0.000411750742932782);
\path [draw=black, semithick]
(axis cs:0,0.000358767841699403)
--(axis cs:0,0.000454357174002696);

\path [draw=black, semithick]
(axis cs:1,4.20684556360682e-05)
--(axis cs:1,4.50690966914082e-05);

\path [draw=black, semithick]
(axis cs:2,2.99361495121354e-05)
--(axis cs:2,3.66959010721678e-05);

\path [draw=black, semithick]
(axis cs:3,0.000367016416771179)
--(axis cs:3,0.000570700356707135);

\path [draw=black, semithick]
(axis cs:4,0.000260461088330896)
--(axis cs:4,0.000405247182058691);

\path [draw=black, semithick]
(axis cs:5,0.00039360207869171)
--(axis cs:5,0.000429899407173853);

\end{axis}

\end{tikzpicture}

%% file: figures/tikz/tissue_manipulation_last_step.tex
% This file was created with tikzplotlib v0.10.1.
\begin{tikzpicture}

\definecolor{crimson2143940}{RGB}{214,39,40}
\definecolor{darkgray176}{RGB}{176,176,176}
\definecolor{forestgreen4416044}{RGB}{44,160,44}
\definecolor{mediumpurple148103189}{RGB}{148,103,189}
\definecolor{steelblue31119180}{RGB}{31,119,180}
\definecolor{darkslategray46}{RGB}{46,46,46}

\begin{axis}[
width=\textwidth,
height=5cm,
log basis y={10},
tick align=outside,
tick pos=left,
x grid style={darkgray176},
xmajorgrids,
xmin=-0.69, xmax=5.69,
xtick style={color=black},
xtick={0,1,2,3,4,5},
xticklabel style={align=center},
xticklabels={
  C=1,
  C=5 \\ \gls{ltsgns},
  C=10,
  \gls{mgn},
  \gls{mgn} \\ (M),
  \gls{mgn} \\ (MP)
},
y grid style={darkgray176},
ylabel={Last step Mean Squared Error},
ymajorgrids,
ymin=0.00001, ymax=0.01,
ymode=log,
ytick style={color=black},
grid=both,
yminorticks=true,
% yticklabels={
%   \(\displaystyle {10^{-6}}\),
%   \(\displaystyle {10^{-5}}\),
%   \(\displaystyle {10^{-4}}\),
%   \(\displaystyle {10^{-3}}\),
%   \(\displaystyle {10^{-2}}\),
%   \(\displaystyle {10^{-1}}\)
% }
]
\draw[draw=none,fill=steelblue31119180] (axis cs:-0.4,0.00000001) rectangle (axis cs:0.4,0.00146979394776281);
\draw[draw=none,fill=steelblue31119180, postaction={pattern=north east lines, pattern color=darkslategray46}] (axis cs:0.6,0.00000001) rectangle (axis cs:1.4,8.75681260367855e-05);
\draw[draw=none,fill=steelblue31119180,  postaction={pattern=north west lines, pattern color=darkslategray46}] (axis cs:1.6,0.00000001) rectangle (axis cs:2.4,6.09537622949574e-05);
\draw[draw=none,fill=forestgreen4416044] (axis cs:2.6,0.00000001) rectangle (axis cs:3.4,0.00132207975548226);
\draw[draw=none,fill=crimson2143940] (axis cs:3.6,0.00000001) rectangle (axis cs:4.4,0.000893655477557331);
\draw[draw=none,fill=mediumpurple148103189] (axis cs:4.6,0.00000001) rectangle (axis cs:5.4,0.00141704721609131);
\path [draw=black, semithick]
(axis cs:0,0.000715964493885555)
--(axis cs:0,0.00222362340164006);

\path [draw=black, semithick]
(axis cs:1,6.6105706066966e-05)
--(axis cs:1,0.000109030546006605);

\path [draw=black, semithick]
(axis cs:2,4.88485500109926e-05)
--(axis cs:2,7.30589745789223e-05);

\path [draw=black, semithick]
(axis cs:3,0.00100631461135001)
--(axis cs:3,0.00163784489961452);

\path [draw=black, semithick]
(axis cs:4,0.000685268207586289)
--(axis cs:4,0.00110204274752837);

\path [draw=black, semithick]
(axis cs:5,0.000686067204686912)
--(axis cs:5,0.0021480272274957);

\end{axis}

\end{tikzpicture}

%% file: main/figure_wrappers/03_figure_qualitative_results.tex
\begin{figure}
    \centering
    \includegraphics[width=\textwidth]{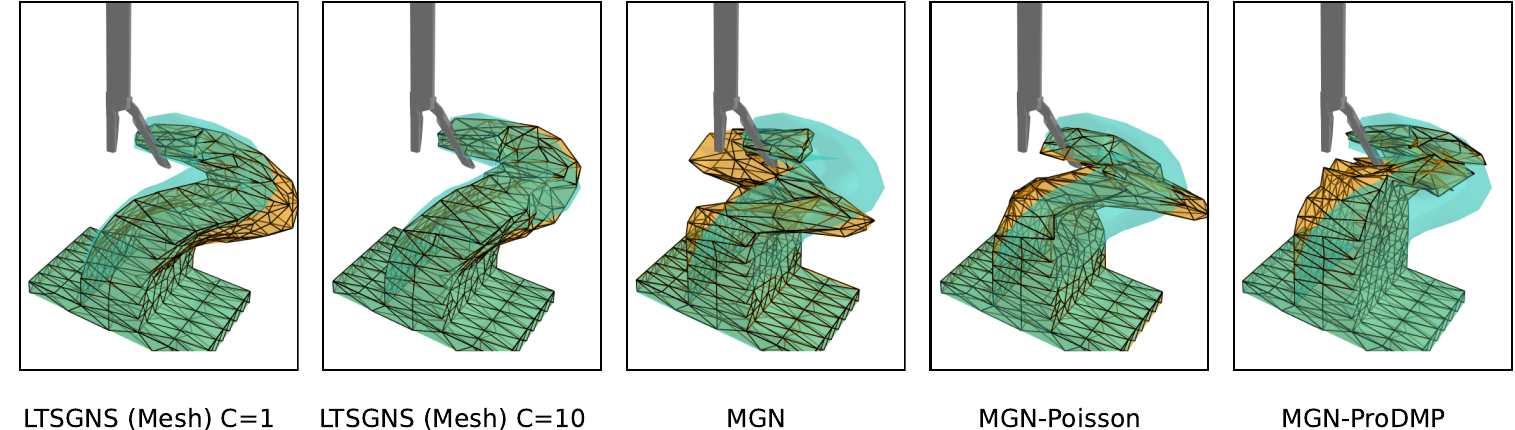}
    \caption{
        Final simulation step of an exemplary test trajectory for the \textit{Tissue Manipulation} task across different methods. 
        Blue denotes the ground position of the deformable object, while the wireframe and yellow shading outline the predicted mesh.
        Only~\gls{ltsgns} provides accurate predictions, which further improve as the size of the context set increases.
    }
    \label{fig:qualitative_results}
\end{figure}

%% file: main/7_conclusion.tex
\section{Conclusion}
We introduce \glsfirst{ltsgns}, a novel Graph Network Simulator that employs meta-learning and movement primitives for accurate probabilistic predictions in physical simulations. 
Our model uses meta-learning and movement primitives to effectively addresses the issue of error accumulation and dynamically adapt to context information during inference.
\gls{ltsgns} is also able to accommodate sensory inputs like point clouds during inference, broadening its applicability in real-world scenarios. 
We experimentally validate the effectiveness of our approach compared to existing baselines, particularly in tasks involving uncertain material properties. 
The model's ability to produce distributions over trajectories adds an extra layer of robustness, making it a valuable tool for both academic research and practical applications where data is sparse.

% We present~\glsfirst{ltsgns}, a novel Graph Network Simulator that utilizes meta-learning and movement primitives for precise probabilistic prediction over simulation trajectories in physical systems.
% Our model adapts to provided context information during inference while mitigating the problem of accumulating errors that related approaches struggle with.
% Furthermore, our approach is versatile enough to incorporate sensory data types, such as point clouds, enhancing its utility in practical applications requiring robust and generalizable graph network simulators. 
% We experimentally validate the effectiveness of our approach, outperforming state-of-the-art graph network simulators on tasks with uncertain material properties,
% Importantly, our probabilistic perspective yields distributions over simulation trajectories, which offers an additional layer of robustness, particularly when dealing with materials whose properties may be uncertain.
% This aspect further solidifies the model's potential for both research and real-world applications.

% We introduce \glsfirst{ltsgns}, a pioneering approach to Graph Network Simulators that leverages meta-learning and movement primitives for precise probabilistic prediction over simulation trajectories in physical systems. Our model is designed to dynamically adapt to context information provided during inference, thereby offering a solution to the problem of error accumulation—a challenge often encountered in existing methods. This feature enables high-fidelity simulations even when only sparse data points are available.

% \par

\paragraph{Limitations and Future Work}
We currently consider each trajectory as a task, and require data from either states or point clouds of this trajectory to fit our model during inference. 
However, as generating such data is often impractical in real-world scenarios, we plan to instead group different material properties into tasks.
This shift will enable the model to encode abstract material behavior, without being tied to a particular mesh topology or simulation.
We further aim to extend our approach to accommodate longer simulations to fully capitalize on the benefits of incorporating movement primitives. 
Combining both, we plan to apply our model to real-world deformations. 
Here, an additional focus will be on integrating online re-planning of trajectories, thereby enhancing prediction accuracy when live sensory information is available.

% We currently view each trajectory as a task, requiring either states or point clouds from this trajectory to fit our model during inference.
% In practical scenarios, generating this data is often not feasible. 
% Instead, we will extend the approach to see a given material property as a task. 
% This allows for the context to encode the behavior of the material under deformation, independent of a concrete mesh topology or simulation.
% We also want to extend our approach to longer simulations to fully utilize the advantages of the movement primitives, and apply it to real-world deformations.
% Here, we want to include online re-planning of the trajectory to get more accurate predictions when given live sensory information.

%% file: appendix/experimental_protocol.tex
\section{Experimental Protocol}
\label{app_sec:experimental_protocol}

In order to promote reproducibility, we offer comprehensive information regarding our experimental methodology. In Table\ref{tab:params}, we give our used hyperparameters for the experiments. To obtain a more detailed description of the tasks and the dataset, please refer to \cite{linkerhagner2023grounding} Appendix B. 

In this paper, we use a training approach that adapts its batch size dynamically based on the context set size. To achieve this, we define a batch cost as an upper bound, calculated as 0.8 plus 0.2 times the batch size. The batch size is then meticulously chosen to align with the specified batch cost, allowing us to optimize our training process effectively. Specifically, for the deformable plate task, we set a target total batch cost of 60, whereas for the tissue manipulation task, we aim for a batch cost of 300. The training for the deformable plate task was conducted on an NVIDIA GeForce RTX 3080 GPU, while the tissue manipulation task utilized an NVIDIA A100 GPU for efficient processing and optimization. The availability of more powerful hardware enabled us to opt for a larger batch cost in the context of the tissue manipulation task.

For each trajectory, we generate 30 auxiliary training tasks, as inspired by the approach outlined in \cite{volpp2023}, Appendix A3.2. In this process, we randomly select pairs of $(x_{l, t}, \bm y_{l, t})$ for each auxiliary training task. Additionally, we designate one of the first 30 time steps as the time steps for constructing the graphs denoted as $\mathcal{G}_l$. 
During the testing phase, the tasks consistently make use of the initial mesh for the corresponding $\mathcal{G}_l$, and we adapt the context size $C$ differently for each evaluation scenario.

\begin{table}[ht]
\centering
\caption{Table listing the hyperparameters and configurations of the experiments}
\begin{tabularx}{.7\textwidth}{l l}
\hline
Parameter & Value \\
\hline
Batches per epoch & $500$ \\
Epochs & $1000$ \\
Node feature dimension & $128$ \\
Latent variable dimension & $8$ \\
Decoder dimension (deformable plate) & $128$ \\
Decoder dimension (tissue manipulation) & $256$ \\
Message passing blocks & 5 \\
Message passing repeats (\gls{mgn}) & 1 \\
Message passing repeats (\gls{ltsgns}) & 2 \\
\gls{mpn} Aggregation function & Mean \\
Learning rate & $5.0e-4$ \\
Gaussian likelihood standard deviation & $0.01$ \\
Number of $z$ samples for ELBO estimation & $32$ \\
Number of \gls{gmm} components & 3 \\
Auxiliary train tasks per trajectory & $30$ \\
Activation function & Leaky ReLU \\
SEMTRUX component KL bound & 0.01 \\
Number of \gls{prodmp} basis functions & $10$ \\
\hline
\end{tabularx}
\label{tab:params}
\end{table}

%% file: appendix/additional_results.tex
\section{Additional Results}
\label{app_sec:additional_results}

\subsection{Evaluations.}
\label{app_ssec:additional_results_quantitative}
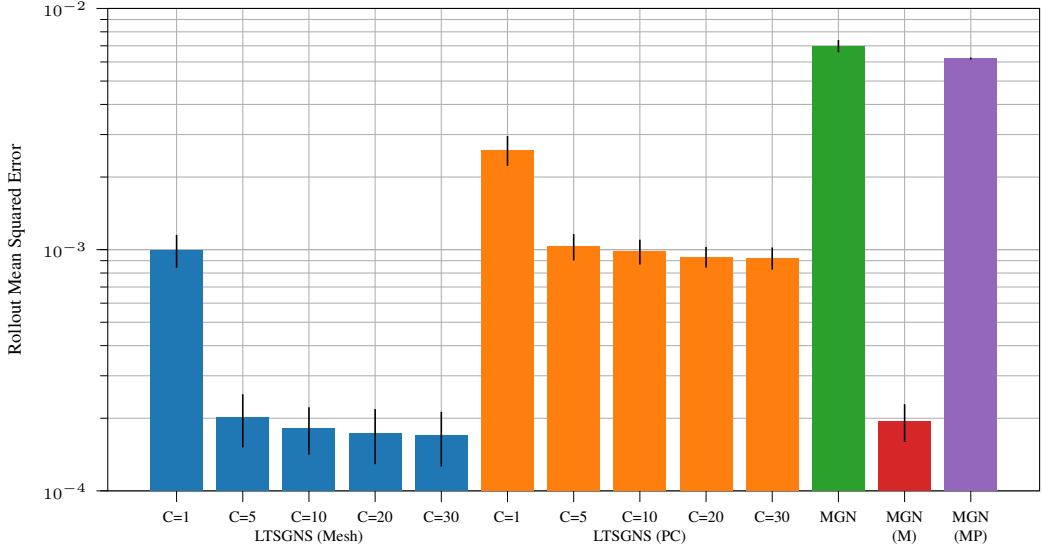
\begin{figure}[t]
    \input{figures/tikz/deformable_plate_all_methods}
    \caption{
    (Left) Rollout and (Right) Last Step \gls{mse} for larger context sizes ($C=20$, $C=30$) compared to the methods presented in Figure~\ref{fig:bar_chart_deformable_plate} for the \textit{Tissue Manipulation} task.
    }
    \label{app_fig:bar_chart_deformable_plate}
\end{figure}

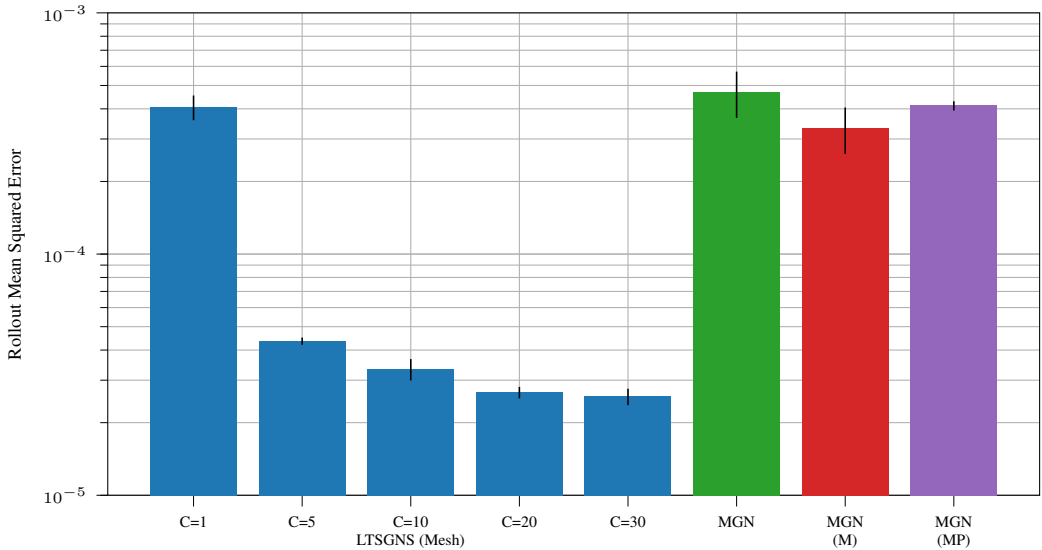
\begin{figure}[t]
    \input{figures/tikz/tissue_manipulation_all_methods}
    \caption{
    (Left) Rollout and (Right) Last Step \gls{mse} for larger context sizes ($C=20$, $C=30$) compared to the methods presented in Figure~\ref{fig:bar_chart_tissue_manipulation} for the \textit{Tissue Manipulation} task.
    }
    \label{app_fig:bar_chart_tissue_manipulation}
\end{figure}

We additionally show how~\gls{ltsgns} performs for larger context sizes in Figures~\ref{app_fig:bar_chart_deformable_plate} (\textit{Deformable Plate}) and~\ref{app_fig:bar_chart_tissue_manipulation} (\textit{Tissue Manipulation}).
While the inclusion of more context information generally enhances performance, it sometimes diminishes performance for the Last Step \gls{mse} when point clouds are used as the context. 
We hypothesize that this effect occurs because a larger context size shifts the balance between the likelihood $L$ and the prior $p(z_l)$, thereby magnifying any existing inaccuracies in the model that may exist in more complex tasks.
% \begin{figure}[t]
%     \centering
%     \begin{subfigure}[b]{0.49\linewidth} % Adjust the width as needed
%         \input{figures/tikz/deformable_plate}
%     \end{subfigure}
%     \hfill % Horizontal spacing between subfigures
%     \begin{subfigure}[b]{0.5\linewidth} % Adjust the width as needed
%         \input{figures/tikz/deformable_plate_last_step}
%     \end{subfigure}
%     \caption{
%     Rollout Last Step \gls{mse} of the different methods on the \textit{Deformable Plate} task. 
%     Both metrics are plotted on a logarithmic scale.
%     \gls{ltsgns} outperforms~\gls{mgn} and its \gls{mp} variant for both metrics from a single context point, further improving its performance when provided with additional context information. 
%     During inference, point cloud information can be used as a context set, making for easier context acquisition at the cost of slightly worse predictions.
%     \todo{we note that the last step is unstable}
%     }

\newpage

\subsection{Visualizations.}
\label{app_ssec:additional_results_qualitative}

\begin{figure}[h]
    \centering
    \includegraphics[width=0.8\textwidth]{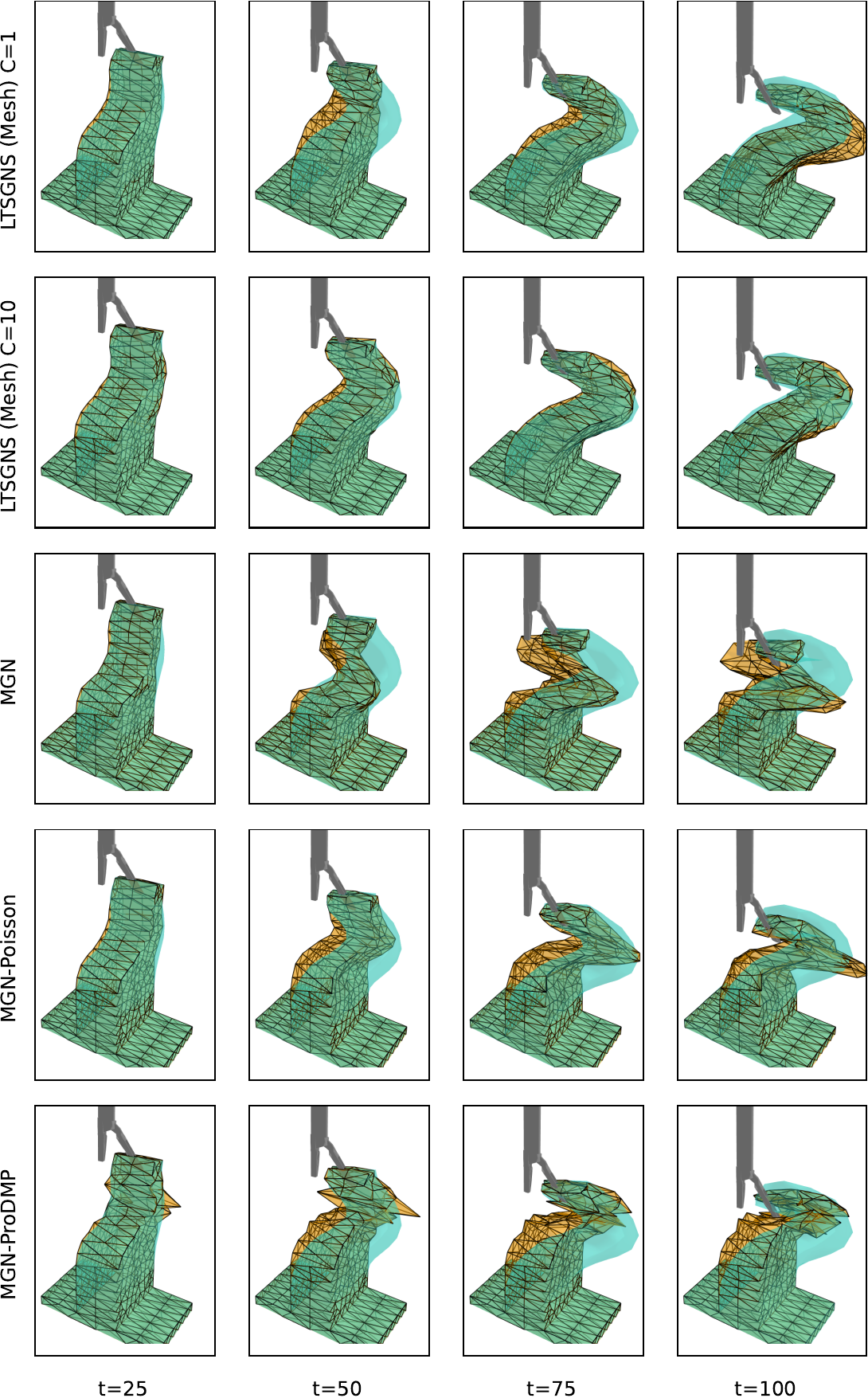}
        \caption{Simulation over time of an exemplary test trajectory of the \textit{Tissue Manipulation} task across different methods. Blue denotes the ground position of the deformable object, while the wireframe and yellow shading outline the predicted mesh.}
    \label{app_fig:tissue_manipulation_appendix}
\end{figure}

We provide additional visualizations for all methods over different timesteps. 
Figure~\ref{app_fig:bar_chart_deformable_plate} shows a test trajectory for the \textit{Deformable Plate} task, and Figure~\ref{app_fig:tissue_manipulation_appendix} visualizes the same for the \textit{Tissue Manipulation} task.
Both figures show that~\gls{ltsgns} performs similar to~\gls{mgn} for a single context point, but significantly improves performance when provided with additional information.
Especially for $C=10$ data points in the context set, the predictions are visually consistent with the ground truth.

\begin{figure}[h]
    \centering
    \includegraphics[width=\textwidth]{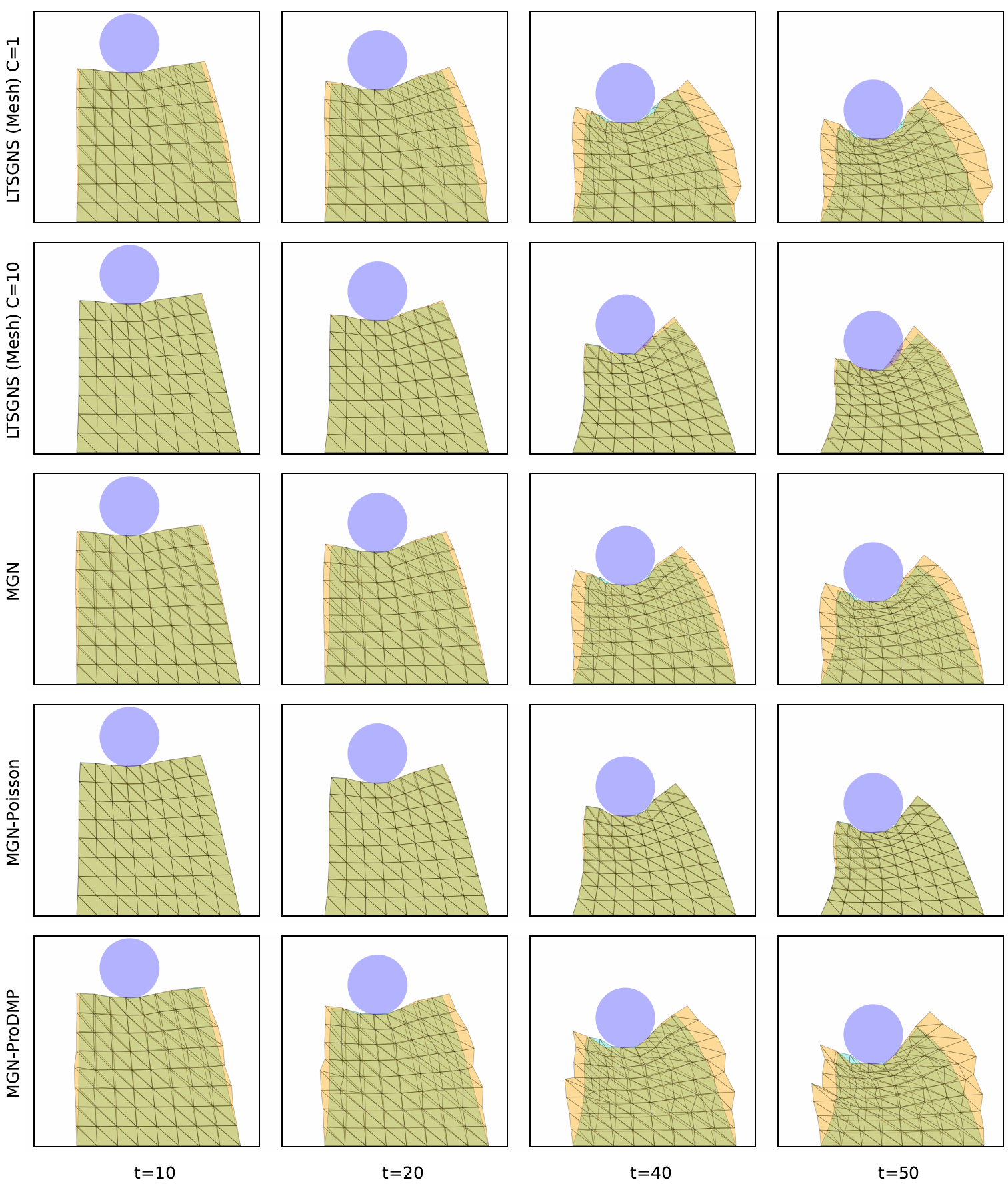}
        \caption{Simulation over time of an exemplary test trajectory of the \textit{Deformable Plate} task across different methods. Blue denotes the ground position of the deformable object, while the wireframe and yellow shading outline the predicted mesh.}
    \label{app_fig:deformable_plate_appendix}
\end{figure}

%% file: figures/tikz/deformable_plate_all_methods.tex
% This file was created with tikzplotlib v0.10.1.
\begin{tikzpicture}

\definecolor{crimson2143940}{RGB}{214,39,40}
\definecolor{darkgray176}{RGB}{176,176,176}
\definecolor{darkorange25512714}{RGB}{255,127,14}
\definecolor{forestgreen4416044}{RGB}{44,160,44}
\definecolor{mediumpurple148103189}{RGB}{148,103,189}
\definecolor{steelblue31119180}{RGB}{31,119,180}

\begin{axis}[
width=\textwidth,
log basis y={10},
tick align=outside,
tick pos=left,
x grid style={darkgray176},
height=8cm,
xmajorgrids,
xmin=-1.04, xmax=13.04,
xtick style={color=black},
xtick={0,1,2,3,4,5,6,7,8,9,10,11,12},
xticklabel style={align=center},
xticklabels={
  C=1,
  C=5,
  C=10 \\ \gls{ltsgns} (Mesh),
  C=20,
  C=30,
  C=1,
  C=5,
  C=10 \\ \gls{ltsgns} (PC),
  C=20,
  C=30,
  \gls{mgn},
  \gls{mgn} \\ (M),
  \gls{mgn} \\ (MP),
},
y grid style={darkgray176},
ylabel={Rollout Mean Squared Error},
ymajorgrids,
ymin=0.0001, ymax=0.01,
ymode=log,
ytick style={color=black},
grid=both,
yminorticks=true,
% yticklabels={
%   \(\displaystyle {10^{-5}}\),
%   \(\displaystyle {10^{-4}}\),
%   \(\displaystyle {10^{-3}}\),
%   \(\displaystyle {10^{-2}}\),
%   \(\displaystyle {10^{-1}}\)
% }
]
\draw[draw=none,fill=steelblue31119180] (axis cs:-0.4,0.00000001) rectangle (axis cs:0.4,0.000995992333628237);
\draw[draw=none,fill=steelblue31119180] (axis cs:0.6,0.00000001) rectangle (axis cs:1.4,0.000201520411064848);
\draw[draw=none,fill=steelblue31119180] (axis cs:1.6,0.00000001) rectangle (axis cs:2.4,0.000181714750942774);
\draw[draw=none,fill=steelblue31119180] (axis cs:2.6,0.00000001) rectangle (axis cs:3.4,0.000173625175375491);
\draw[draw=none,fill=steelblue31119180] (axis cs:3.6,0.00000001) rectangle (axis cs:4.4,0.000169429220841266);
\draw[draw=none,fill=darkorange25512714] (axis cs:4.6,0.00000001) rectangle (axis cs:5.4,0.00259234984405339);
\draw[draw=none,fill=darkorange25512714] (axis cs:5.6,0.00000001) rectangle (axis cs:6.4,0.00103087805910036);
\draw[draw=none,fill=darkorange25512714] (axis cs:6.6,0.00000001) rectangle (axis cs:7.4,0.000983766315039247);
\draw[draw=none,fill=darkorange25512714] (axis cs:7.6,0.00000001) rectangle (axis cs:8.4,0.000934745697304606);
\draw[draw=none,fill=darkorange25512714] (axis cs:8.6,0.00000001) rectangle (axis cs:9.4,0.000923759874422103);
\draw[draw=none,fill=forestgreen4416044] (axis cs:9.6,0.00000001) rectangle (axis cs:10.4,0.00698697278276086);
\draw[draw=none,fill=crimson2143940] (axis cs:10.6,0.00000001) rectangle (axis cs:11.4,0.00019408235384617);
\draw[draw=none,fill=mediumpurple148103189] (axis cs:11.6,0.00000001) rectangle (axis cs:12.4,0.00620234319940209);
\path [draw=black, semithick]
(axis cs:0,0.000841293610898635)
--(axis cs:0,0.00115069105635784);

\path [draw=black, semithick]
(axis cs:1,0.000151577272235135)
--(axis cs:1,0.000251463549894561);

\path [draw=black, semithick]
(axis cs:2,0.000141369666997861)
--(axis cs:2,0.000222059834887688);

\path [draw=black, semithick]
(axis cs:3,0.000129039177429837)
--(axis cs:3,0.000218211173321146);

\path [draw=black, semithick]
(axis cs:4,0.000126127912459664)
--(axis cs:4,0.000212730529222869);

\path [draw=black, semithick]
(axis cs:5,0.00222597537485019)
--(axis cs:5,0.00295872431325658);

\path [draw=black, semithick]
(axis cs:6,0.00090185292998838)
--(axis cs:6,0.00115990318821234);

\path [draw=black, semithick]
(axis cs:7,0.000867873934405715)
--(axis cs:7,0.00109965869567278);

\path [draw=black, semithick]
(axis cs:8,0.000842965760204893)
--(axis cs:8,0.00102652563440432);

\path [draw=black, semithick]
(axis cs:9,0.000826344372704707)
--(axis cs:9,0.0010211753761395);

\path [draw=black, semithick]
(axis cs:10,0.00657996642208122)
--(axis cs:10,0.0073939791434405);

\path [draw=black, semithick]
(axis cs:11,0.000159421332010872)
--(axis cs:11,0.000228743375681468);

\path [draw=black, semithick]
(axis cs:12,0.00614262091360138)
--(axis cs:12,0.00626206548520281);

\end{axis}

\end{tikzpicture}

%% file: figures/tikz/tissue_manipulation_all_methods.tex
% This file was created with tikzplotlib v0.10.1.
\begin{tikzpicture}

\definecolor{crimson2143940}{RGB}{214,39,40}
\definecolor{darkgray176}{RGB}{176,176,176}
\definecolor{forestgreen4416044}{RGB}{44,160,44}
\definecolor{mediumpurple148103189}{RGB}{148,103,189}
\definecolor{steelblue31119180}{RGB}{31,119,180}

\begin{axis}[
width=\textwidth,
log basis y={10},
tick align=outside,
tick pos=left,
x grid style={darkgray176},
xmajorgrids,
height=8cm,
xmin=-0.79, xmax=7.79,
xtick style={color=black},
xtick={0,1,2,3,4,5,6,7},
xticklabel style={align=center},
xticklabels={
  C=1,
  C=5, 
  C=10 \\ \gls{ltsgns} (Mesh),
  C=20, 
  C=30, 
  \gls{mgn},
  \gls{mgn} \\ (M),
  \gls{mgn} \\ (MP)
},
y grid style={darkgray176},
ylabel={Rollout Mean Squared Error},
ymajorgrids,
ymin=0.00001, ymax=0.001,
ymode=log,
ytick style={color=black},
grid=both,
yminorticks=true,
% yticklabels={
%   \(\displaystyle {10^{-6}}\),
%   \(\displaystyle {10^{-5}}\),
%   \(\displaystyle {10^{-4}}\),
%   \(\displaystyle {10^{-3}}\),
%   \(\displaystyle {10^{-2}}\)
% }
]
\draw[draw=none,fill=steelblue31119180] (axis cs:-0.4,0.00000001) rectangle (axis cs:0.4,0.000406562507851049);
\draw[draw=none,fill=steelblue31119180] (axis cs:0.6,0.00000001) rectangle (axis cs:1.4,4.35687761637382e-05);
\draw[draw=none,fill=steelblue31119180] (axis cs:1.6,0.00000001) rectangle (axis cs:2.4,3.33160252921516e-05);
\draw[draw=none,fill=steelblue31119180] (axis cs:2.6,0.00000001) rectangle (axis cs:3.4,2.66793176706415e-05);
\draw[draw=none,fill=steelblue31119180] (axis cs:3.6,0.00000001) rectangle (axis cs:4.4,2.56415041803848e-05);
\draw[draw=none,fill=forestgreen4416044] (axis cs:4.6,0.00000001) rectangle (axis cs:5.4,0.000468858386739157);
\draw[draw=none,fill=crimson2143940] (axis cs:5.6,0.00000001) rectangle (axis cs:6.4,0.000332854135194793);
\draw[draw=none,fill=mediumpurple148103189] (axis cs:6.6,0.00000001) rectangle (axis cs:7.4,0.000411750742932782);
\path [draw=black, semithick]
(axis cs:0,0.000358767841699403)
--(axis cs:0,0.000454357174002696);

\path [draw=black, semithick]
(axis cs:1,4.20684556360682e-05)
--(axis cs:1,4.50690966914082e-05);

\path [draw=black, semithick]
(axis cs:2,2.99361495121354e-05)
--(axis cs:2,3.66959010721678e-05);

\path [draw=black, semithick]
(axis cs:3,2.52234627730547e-05)
--(axis cs:3,2.81351725682282e-05);

\path [draw=black, semithick]
(axis cs:4,2.36367418743416e-05)
--(axis cs:4,2.7646266486428e-05);

\path [draw=black, semithick]
(axis cs:5,0.000367016416771179)
--(axis cs:5,0.000570700356707135);

\path [draw=black, semithick]
(axis cs:6,0.000260461088330896)
--(axis cs:6,0.000405247182058691);

\path [draw=black, semithick]
(axis cs:7,0.00039360207869171)
--(axis cs:7,0.000429899407173853);

\end{axis}

\end{tikzpicture}